\documentclass[letterpaper, 10 pt, conference]{ieeeconf}  

\IEEEoverridecommandlockouts                              

\usepackage{graphics} 
\usepackage{epsfig} 
\usepackage{mathptmx} 
\usepackage{times} 
\usepackage{amsmath} 
\usepackage{amssymb}  
\usepackage{booktabs}   
\usepackage{tabularx}   
\usepackage{array}      
\usepackage{multirow}
\usepackage{url}

\title{\LARGE \bf
Data-Driven Optimization of EV Charging Station Placement Using Causal Discovery
}

\author{Julius Stephan Junker$^{1}$, Rong Hu$^{2}$, Ziyue Li$^{1}$ and Wolfgang Ketter$^{1}$
\thanks{$^{1}$The authors are with the Department of Information Systems, University of Cologne, Cologne, Germany
        {\tt\small \{jjunker2@smail, zlibn@wiso, ketter@wiso\}.uni-koel\newline n.de}}%
\thanks{$^{2}$The author is with the College of Computer Science and Electronic Engineering, Hunan University, Changsha, China
        {\tt\small upupwords@hnu.edu.cn}}%
}

\begin{document}

\maketitle
\thispagestyle{empty}
\pagestyle{empty}

\begin{abstract}
This paper addresses the critical challenge of optimizing electric vehicle charging station placement through a novel data-driven methodology employing causal discovery techniques. While traditional approaches prioritize economic factors or power grid constraints, they often neglect empirical charging patterns that ultimately determine station utilization. We analyze extensive charging data from Palo Alto and Boulder (337,344 events across 100 stations) to uncover latent relationships between station characteristics and utilization. Applying structural learning algorithms (NOTEARS and DAGMA) to this data reveals that charging demand is primarily determined by three factors: proximity to amenities, EV registration density, and adjacency to high-traffic routes. These findings, consistent across multiple algorithms and urban contexts, challenge conventional infrastructure distribution strategies. We develop an optimization framework that translates these insights into actionable placement recommendations, identifying locations likely to experience high utilization based on the discovered dependency structures. The resulting site selection model prioritizes strategic clustering in high-amenity areas with substantial EV populations rather than uniform spatial distribution. Our approach contributes a framework that integrates empirical charging behavior into infrastructure planning, potentially enhancing both station utilization and user convenience. By focusing on data-driven insights instead of theoretical distribution models, we provide a more effective strategy for expanding charging networks that can adjust to various stages of EV market development.
\end{abstract}


\section{INTRODUCTION}

The electrification of transportation represents one of the most significant transformations in modern mobility systems, with electric vehicles (EVs) emerging as a critical solution to reduce greenhouse gas emissions and dependence on fossil fuels \cite{Singh2020}. However, widespread EV adoption depends on sufficient charging infrastructure, which remains a major barrier \cite{Aljaidi2022, Yang2022}. As the EV market continues to grow, the strategic placement of charging stations has become increasingly important, requiring methodologies that balance technical feasibility, economic viability, and user convenience  \cite{Banegas2023}.

Traditional approaches to electric vehicle charging station (EVCS) placement have primarily focused on economic optimization from the perspective of charging network operators or technical integration with power distribution networks \cite{Yang2022}. While these considerations are undoubtedly important, they often neglect the empirical patterns of user behavior and demand that ultimately determine a charging station's utilization and effectiveness \cite{Adenaw2022}. As Singh \emph{et al.} \cite{Singh2020} established in their review of 211 peer-reviewed papers, charging infrastructure and policy-making are the most significant factors influencing EV adoption, highlighting the critical importance of well-designed charging networks.

The literature on EVCS placement highlights three primary methodological approaches. Meta-heuristic methods offer computational efficiency and can handle conflicting objectives but often settle in local solutions. Multi-criteria decision-making (MCDM) approaches incorporate multiple factors and are widely applicable, yet they suffer from subjectivity and difficulty in achieving globally optimal solutions \cite{Yang2022}. Analytical methods provide high precision but require accurate system models and significant computational resources \cite{Aljaidi2022}. Additionally, Geographic Information System (GIS)-based approaches have gained prominence, with \cite{Banegas2023} identifying 74 studies employing weighted overlays of spatial factors to determine optimal locations \cite{ye2024survey,liu2024spatial}.

Despite this methodological diversity, a critical gap remains: the limited integration of empirical charging behavior data into location optimization models. While economic considerations such as cost minimization and technical constraints like grid capacity have been extensively studied \cite{Yang2022}, the actual patterns of charging station utilization—and the factors that influence these patterns—are rarely incorporated into optimization frameworks. This gap is particularly notable given the increasing availability of charging event data and the growing understanding that user behavior significantly impacts infrastructure effectiveness \cite{Wolbertus2021}.

Some recent studies have started addressing this issue. Straka \emph{et al.} \cite{Straka2020} analyzed charging data from the Netherlands, identifying correlations between station usage and factors such as proximity to points of interest. Similarly, Wolbertus \emph{et al.} \cite{Wolbertus2021} employed agent-based simulations informed by real-world charging data from Amsterdam to evaluate different deployment strategies. These studies represent important steps toward evidence-based infrastructure planning but have not yet developed comprehensive methodologies that integrate causal insights about charging behavior into optimization frameworks.

Our research bridges this critical gap by developing a data-driven methodology that integrates empirical charging data with causal discovery techniques to optimize charging station placement. Specifically, we:
\begin{itemize}
    \item analyze extensive real-world charging data from two cities (Palo Alto and Boulder) to identify patterns in charging station utilization and their relationship to spatial and contextual factors.
    \item apply structural learning algorithms (NOTEARS and DAGMA) to infer potential causal relationships between station characteristics and charging demand, providing insights beyond mere correlations.
    \item develop an optimization framework that leverages these discovered relationships to identify optimal locations for new charging stations, balancing user convenience with operational considerations.
\end{itemize}
By uncovering the latent dependency structures in charging behavior, we offer a methodologically rigorous approach to charging infrastructure planning that prioritizes actual user patterns rather than theoretical assumptions. The integration of causal discovery techniques represents a novel contribution to the EVCS literature, offering deeper insights into the factors driving charging station utilization and providing a more empirically grounded approach to infrastructure planning.

The remainder of this paper is organized as follows: Section \ref{sec:data} describes the datasets and feature engineering process. Section\ref{sec:meth} presents our methodological framework, including the structural learning approach for demand estimation and the optimization model for site selection. Section \ref{sec:res} presents and discusses our results, highlighting empirical patterns, structural learning insights, and optimization outcomes. Finally, Section \ref{sec:con} concludes with key findings and implications for charging infrastructure planning.

\section{Data And Feature Engineering}\label{sec:data}
\subsection{Dataset Sources and Preprocessing}
Electric vehicle charging behavior is inherently spatial and temporal, necessitating high-quality empirical data to assess the factors influencing charging station utilization. This study utilizes real-world charging event data from two distinct urban environments: Palo Alto, California and Boulder, Colorado. These datasets provide comprehensive records of charging activities across multiple years, enabling robust analysis of usage patterns and their determinants.
Table \ref{tab:metadataChargingEvents} compares charging event datasets from Palo Alto and Boulder, detailing key statistics such as the number of charging events, charging stations, and locations, as well as the observation period. The Palo Alto dataset encompasses a longer historical period (2011-2020), while the Boulder dataset provides more recent observations (2018-2023). Individual charging events contain recorded energy consumption (kWh), precise timestamps, and station location identifiers, forming the foundation for our analysis.

\begin{figure}[htbp]
\begin{center}
\includegraphics[scale=0.53]{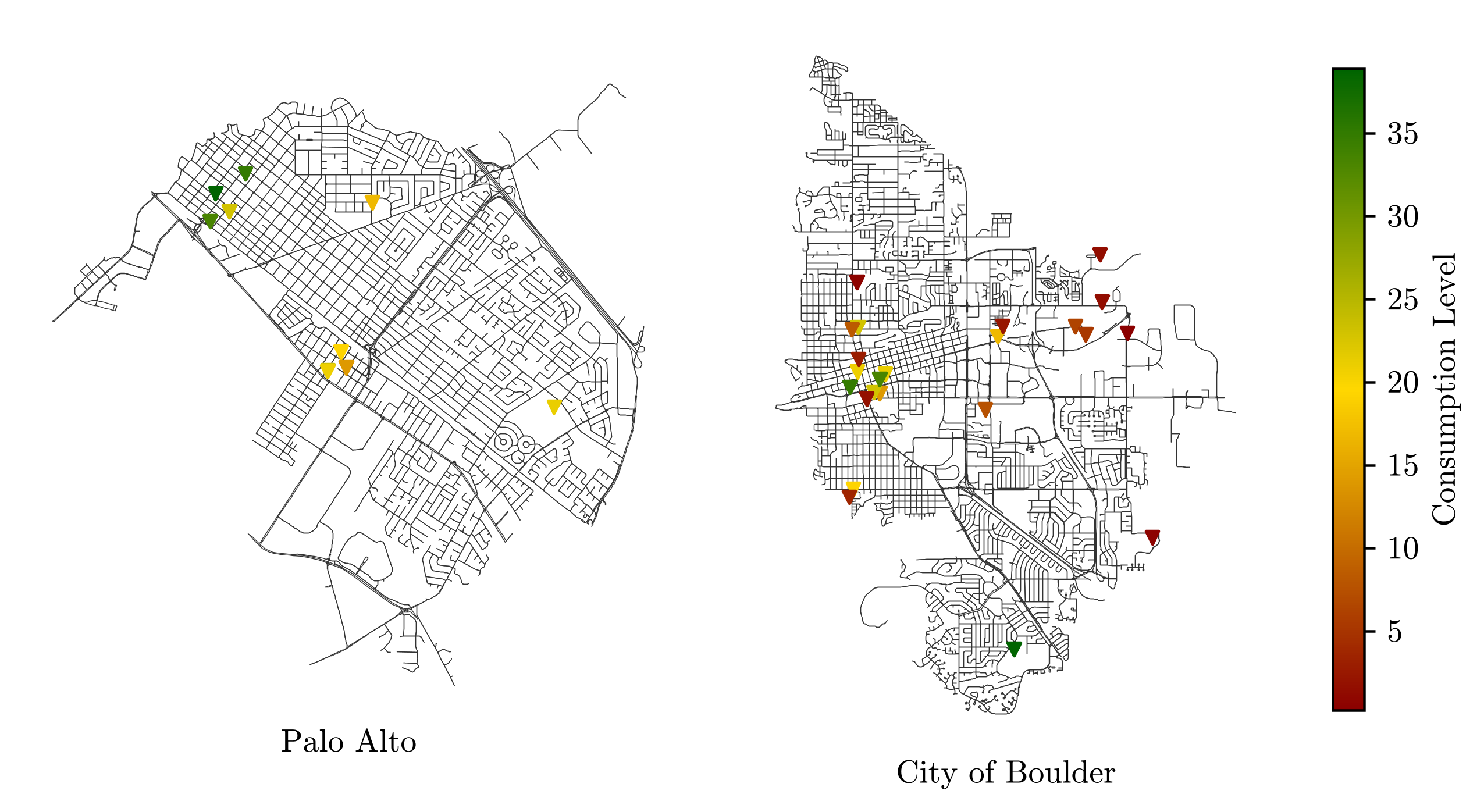}
\caption{CS locations with charging consumption level in Palo Alto and Boulder.}
\label{fig:maps}
\end{center}
\end{figure}

To account for variations in EV adoption over time, preventing newer charging stations from appearing to have inherently higher demand due to increased market penetration, we design a normalized metric of charging consumption level through the following three-step process:
\begin{enumerate}
    \item Adjusted Energy Consumption Per Event. To correct for changes in EV adoption over time, the energy consumption for each charging event is scaled by the ratio of EVs per charging station in a reference year (2011 for Palo Alto and 2018 for Boulder):
    \begin{equation}
    E'_{t} = E_{t} \times \frac{\text{PEV}_{2011} / \text{CS}_{2011}}{\text{PEV}_{t} / \text{CS}_{t}}
    \end{equation}
    where \( E_{t} \) is the raw energy consumption, and \( \text{PEV}_{t} \) and \( \text{CS}_{t} \) represent the number of EVs and charging stations at time \( t \).
    
    \item Average Daily Consumption Per Station. To calculate the average daily energy consumption for each charging station: 
    \begin{equation}
    C_{\text{CS}} = \frac{\sum E'_{t}}{D_{\text{CS}}}
    \end{equation}
    where \( D_{\text{CS}} \) is the total number of days the station has been in operation.

    \item Total Consumption Per Location. To compute the total consumption level for each charging location, we sum the average daily consumption across all stations at that location
    \begin{equation}
    C_{\text{Location}} = \sum_{\text{CS} \in \text{Location}} C_{\text{CS}}
    \end{equation}

\end{enumerate}
This normalization methodology ensures a fair comparison of charging demand across different locations and time periods by accounting for the evolving ratio of EVs to charging stations. The resulting consumption level metric serves as our primary dependent variable in subsequent analyses.
Fig.\ref{fig:maps} shows the CS locations within Palo Alto (left) and Boulder (right) color-coded according to their respective consumption levels, with green indicating high levels and red indicating low levels.

\begin{table}[]
\caption{Metadata-comparison for charging event datasets from Palo Alto and Boulder.}
\label{tab:metadataChargingEvents}
\resizebox{\linewidth}{!}{%
\begin{tabular}{lll}
\hline
                   & Palo Alto\cite{BoulderChargingData}              & Boulder\cite{PaloAltoChargingData}                 \\ \hline
Charging Events    & 259,407                 & 77,937                  \\
Charging Stations  & 48                      & 52                      \\
Locations          & 10                      & 25                      \\
Observation Period & 29/07/2011 -- 31/12/2020 & 01/01/2018 -- 30/11/2023 \\ \hline
\end{tabular}%
}
\end{table}

\begin{table*}[htbp]
\caption{Feature Engineering and Selection}
\label{tab:feature_engineering}
\centering
\renewcommand{\arraystretch}{1.2}
\begin{tabular}{p{4.5cm}  p{5cm}  p{7cm}} 
\toprule
\textbf{Feature Category} & \textbf{Feature Name} & \textbf{Description} \\
\midrule
\multirow{3}{*}{\textbf{Demand-Driven Variables}}  
& \textbf{Traffic Flow Proximity} & Binary indicator (1 = within 70m of an arterial road); assesses station visibility and convenience. \\ \cline{2-3} 
& \textbf{Network Centrality} & Closeness centrality of the station within the road network, computed using OpenStreetMap graphs. \\ \cline{2-3} 
& \textbf{EV Density in Zip Code Area} & Number of registered EVs per zip code, serving as a proxy for latent charging demand. \\ 
\midrule
\multirow{3}{*}{\textbf{Environmental \& Urban Context}}  
& \textbf{Proximity to POIs} & Distance to categorized points of interest (amenities, retail, transport hubs, entertainment, medical). \\ \cline{2-3}
& \textbf{Accommodation Share} & Percentage of buildings designated for overnight stays (hotels, residential complexes). \\ \cline{2-3}
& \textbf{Public Transport Accessibility} & Number of public transport stations within a 400m walkable radius, indicating multimodal accessibility. \\ 
\midrule
\multirow{2}{*}{\textbf{Charging Infrastructure Variables}}  
& \textbf{Number of Chargers per Location} & Total count of charging plugs at each site, reflecting available supply capacity. \\ \cline{2-3}
& \textbf{Distance to Nearest Competing Station} & Road network distance (meters) to the closest alternative charging station, assessing competition effects. \\ 
\bottomrule
\end{tabular}
\end{table*}

\subsection{Feature Engineering and Selection}
To enhance the analytical scope of our investigation, we integrated geospatial data on transportation networks, urban infrastructure, and points of interest (POIs) from multiple sources \cite{California2024EVCount,Colorado2024EVCount,PaloAltoCSData1,BoulderCSData1} with our charging station data. This integration enabled the development of a comprehensive feature set capturing the spatial and contextual factors hypothesized to influence charging behavior.

\begin{figure}[htbp]
    \centering
    \includegraphics[scale=0.6]{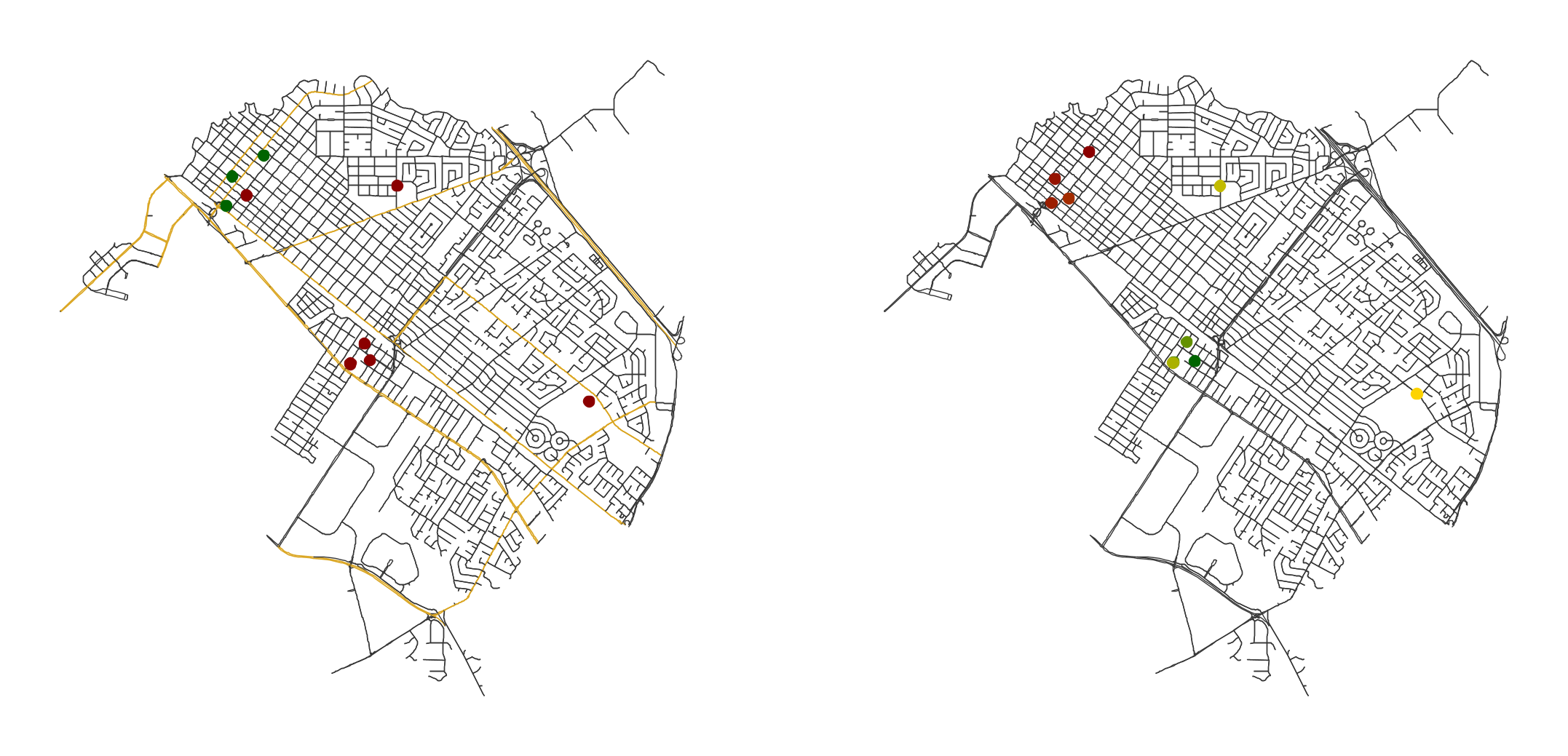}
    \caption{Visualizations for \textit{traffic flow proximity} (left) and \textit{network centrality} (right) exemplary for Palo Alto. In the left map the highlighted streets mark traffic arteries. CS locations within 70 meters straight line distance of a traffic artery are marked green, and CS locations further than that are marked red. In the right plot green (red) stands for a high (low) \textit{network centrality}.}
    \label{fig:ht_nc_plot}
    
\end{figure}

Table \ref{tab:feature_engineering} categorizes these features into demand-driven, environmental, and charging infrastructure variables, selected based on prior research in EV adoption, transportation networks, and urban planning. Factors such as traffic flow, network centrality, and EV density, driven by demand, influence station usage by capturing accessibility and market size. Fig.\ref{fig:ht_nc_plot} visualizes traffic flow proximity and network centrality in Palo Alto. Environmental variables, including POI proximity, accommodation share, and public transport accessibility, reflect location attractiveness and multimodal connectivity. Infrastructure features, such as the number of chargers per location and distance to the nearest competing station, contextualize station performance within the existing network.

These features provide a structured foundation for demand estimation and optimal site selection, supporting subsequent structural learning and optimization models.

\section{Methodology}\label{sec:meth}
This research implements a two-stage methodological framework to address the challenge of optimal EVCS placement. The first stage employs directed acyclic graphs (DAGs) to infer structural relationships between station characteristics and charging demand. The second stage leverages these insights to develop an optimization model for strategic site selection.

\subsection{Structural Learning for Demand Estimation}

The analytical foundation of this research lies in uncovering latent dependency structures among the variables described in Section \ref{sec:data}. DAGs offer a powerful framework for representing these relationships by encoding conditional independence properties within multivariate systems, which has been applies into smart transportation \cite{lan2023mm,lin2023dynamic,lan2024multifun,kai2024deep,peter2024deep}. The identification of such structures from observational data presents a computational challenge characterized as NP-hard. To address this challenge, this study employs two contemporary structural learning algorithms: NOTEARS \cite{Zheng2018} and DAGMA \cite{Bello2022}.

NOTEARS reformulates the traditionally combinatorial DAG learning problem as a continuous optimization task:
\begin{equation}
    \begin{aligned}
        \min_{W \in \mathbb{R}^{d \times d}} \quad & \quad F(W) \\
        \text{subject to } \quad & \quad h(W) = 0,
    \end{aligned}
\end{equation}
where $d$ represents the dimensionality of the variable space, $W$ denotes a weighted adjacency matrix encoding the graph structure, $F: \mathbb{R}^{d \times d} \rightarrow \mathbb{R}$ is a score function capturing goodness-of-fit, and $h: \mathbb{R}^{d \times d} \rightarrow \mathbb{R}$ is a differentiable constraint function ensuring acyclicity.

Given a data matrix $\mathbf{X} \in \mathbb{R}^{n \times d}$ of $n$ observations, NOTEARS minimizes a least-squares score function with $L_1$ regularization:
\begin{equation}
    F(W) = \ell(W;\textbf{X}) + \lambda \|W\|_1 = \frac{1}{2n} \|\mathbf{X} - \mathbf{X}W\|^2_F + \lambda \|W\|_1,
\end{equation}
where $\lambda \geq 0$ is the regularization parameter controlling model sparsity.

The acyclicity constraint is formulated as:
\begin{equation}
    h(W) = \text{trace} (e^{W \circ W}) - d = 0,
\end{equation}
where $e^A$ represents the matrix exponential and $A \circ B$ denotes the Hadamard product.

DAGMA extends this approach through an alternative formulation of acyclicity based on M-matrices. DAGMA characterizes acyclicity using the log-determinant function:
\begin{equation}
    h^s_{\text{ldet}}(W) = - \text{log det}(sI - W \circ W) + d \text{ log } s,
\end{equation}
where $s>0$ is a scalar parameter and $W$ belongs to:
\begin{equation}
    \mathbb{W}^s = \{W \in \mathbb{R}^{d \times d} \mid s> \rho(W \circ W) \},
\end{equation}
with $\rho(A)$ denoting the spectral radius of matrix $A$.


\begin{figure}[htbp]
\begin{center}
\includegraphics[scale=0.54]{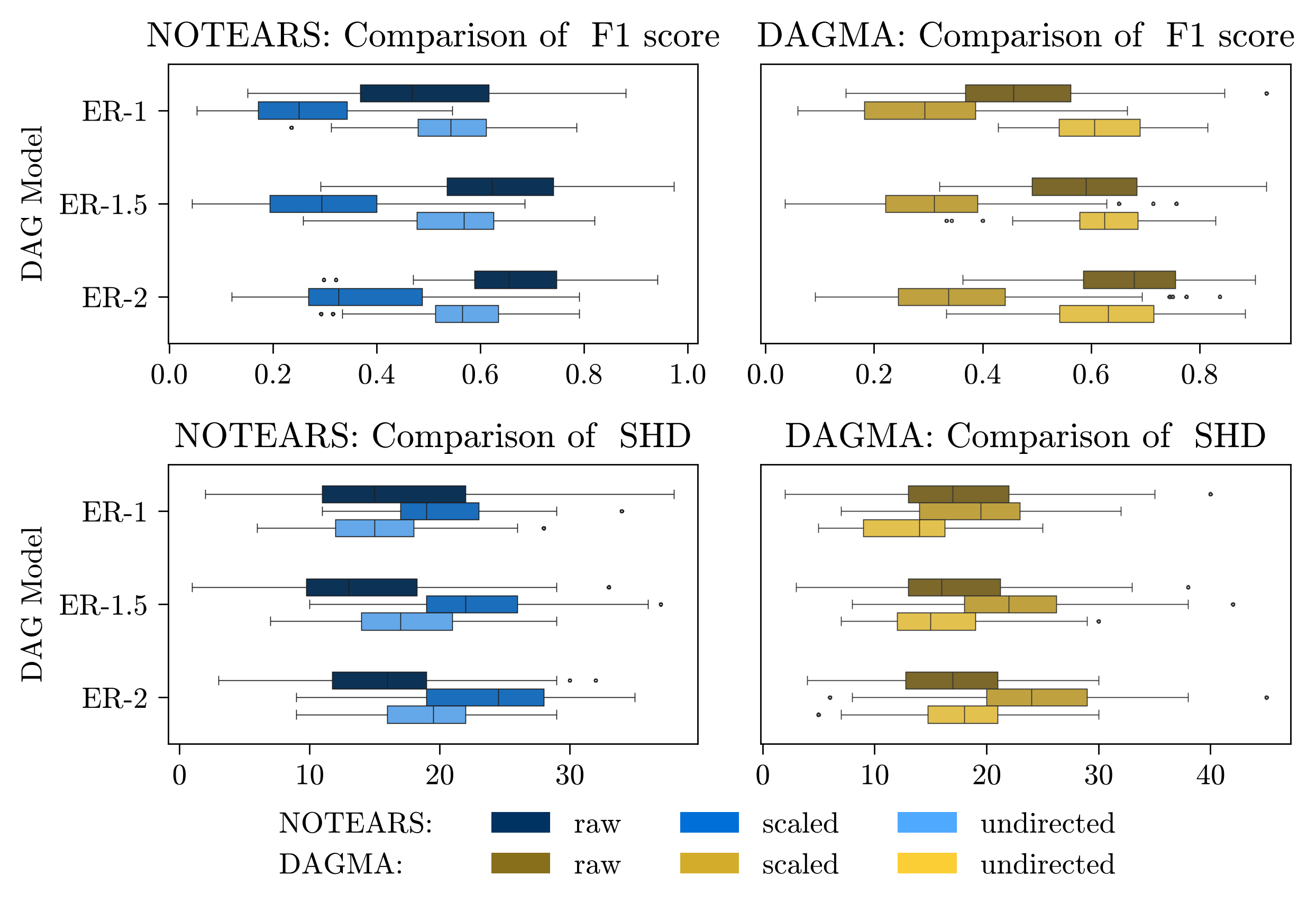}
\caption{Comparison of performance of NOTEARS and DAGMA on raw vs. scaled data. For the case of the scaled data $W_{est}$ is interpreted as defining a DAG as well as defining an undirected graph.}
\label{fig:raw_scaled_undirected}
\end{center}
\end{figure}

The primary output of both algorithms is an estimated weighted adjacency matrix, $W_{est}$, where a non-zero entry $W_{est_{ij}}$indicates a directed edge from variable $i$ to variable $j$. This matrix encodes the inferred dependency structure among the features described in Section \ref{sec:data}, including demand-driven variables, environmental and urban context variables, and charging infrastructure variables.

Understanding and applying $W_{est}$ requires careful consideration of methodological limitations. When used directly, $W_{est}$ represents a DAG where edges suggest directional influences between variables. However, from observational data alone, we can only identify Markov equivalence classes—sets of DAGs that encode the same conditional independence relations but may differ in edge directions for some variable pairs.

Given this limitation, we can interpret $W_{est}$ in two ways: either as a directed graph (accepting the edge directions despite their uncertainty) or as an undirected graph (considering only whether variables are related, ignoring the proposed direction). This dual interpretation acknowledges the inherent limitations of causal discovery from observational data while maximizing the practical utility of the identified structural relationships. Fig. \ref{fig:raw_scaled_undirected} compares these two interpretation approaches across both raw and standardized data, showing how each performs in recovering true underlying structures.

Several methodological considerations inform our structural learning approach:
\begin{itemize}
    \item Non-convex optimization requires careful initialization to avoid local optima in identifying charging demand dependencies.
    \item Identified relationships represent statistical associations requiring domain validation before inferring causality.
    \item Observational data can only identify Markov equivalence classes, not definitive causal directions between urban features and charging utilization.
    \item Feature standardization prevents high-magnitude variables from dominating structural learning.
    \item Feature scale heterogeneity (e.g., EV count vs. binary indicators) necessitates standardization to prevent dominance by magnitude rather than explanatory power.
    \item Variance differentials\cite{Reisach2021} between features may bias directional inferences.
\end{itemize}
These considerations guide our interpretation of the discovered dependency structures between consumption levels and predictor variables, establishing a methodologically sound foundation for the subsequent optimization framework.
\begin{figure*}[htbp]
\begin{center}
\includegraphics[scale=0.4]{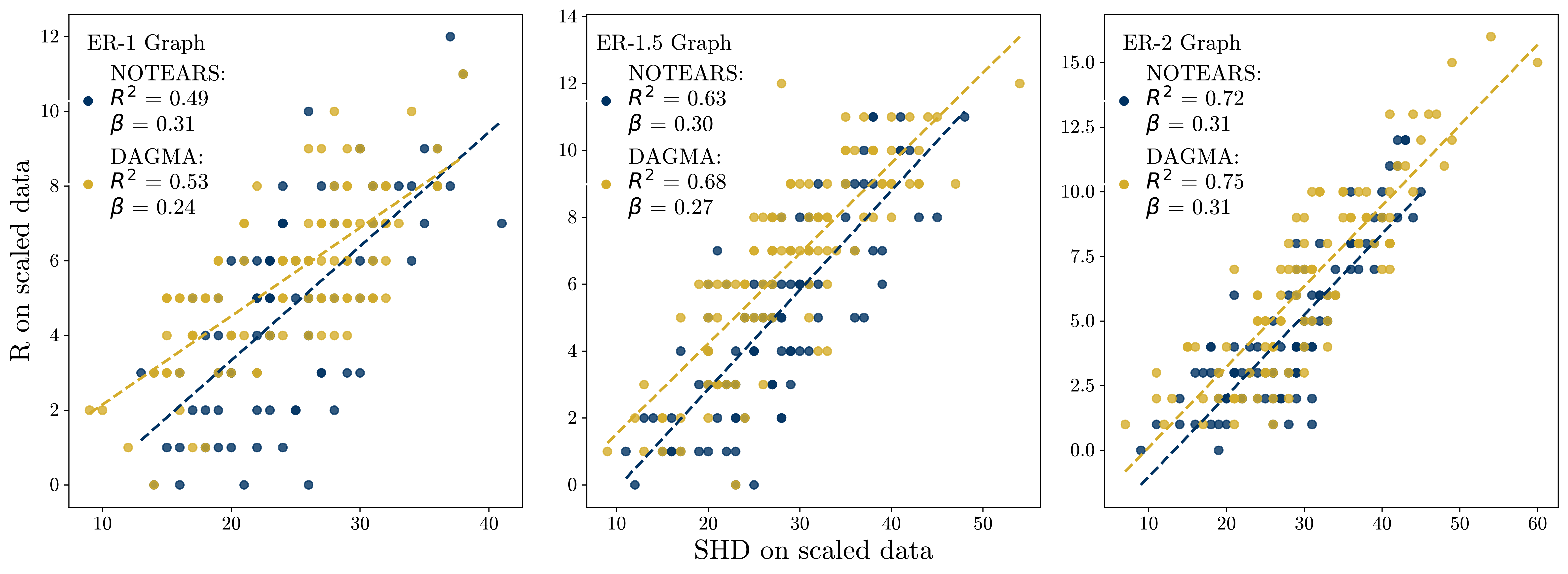}
\caption{SHD vs. the number of reversed edges on scaled data(R for reversed).}
\label{fig:SHD_vs_R}
\end{center}
\end{figure*}

\subsection{Optimization Model for Site Selection}
The second methodological stage leverages insights from structural learning to formulate a principled optimization model for EVCS site selection. Based on the structural analysis of $W_{est}$, a subset of features $F$ that demonstrate significant explanatory power for charging station utilization patterns is selected. These features serve as predictors in a Bayesian regression model:
\begin{equation}
\begin{aligned}
    \textit{consumption\_level}_i &\sim \text{Normal}(\mu_i,\sigma)\\
    \mu_i &= \alpha + \sum_{f \in F} \beta_f \cdot f
\end{aligned}
\end{equation}
The Bayesian framework is particularly appropriate given the limited spatial sample (35 charging locations across both cities), as it enables the incorporation of parameter uncertainty and mitigates overfitting concerns through principled regularization via informative priors.

For each node $v$ in the street network of the study area $P=(V,E)$, a demand score $s_v$ is calculated based on the regression coefficients:
\begin{align}
    s_v = \sum_{f \in F} b_f \cdot f_v,
\end{align}
where $b_f$ represents the posterior mean of parameter $\beta_f$ derived from Markov Chain Monte Carlo sampling (2000 posterior draws), and $f_v$ is the value of feature $f$ at node $v$.

The optimization problem is then formulated as follows:
\begin{align}
    \max \sum_{v \in C} x_v \cdot s_v
\end{align}
where $x_v$ is a binary decision variable indicating whether a charging station is placed at node $v$, and $C$ represents the set of candidate locations defined as:
\begin{equation}
\begin{aligned}
    C = \{ & v \in V \; \mid \;  \forall cs_{rival} \in CS_{rival},\; \text{dist}(v, cs_{rival}) > d_{rival} \text{ and } \\
    & \forall cs_{operator} \in CS_{operator},\; \text{dist}(v, cs_{operator}) > d_{operator} \}.
\end{aligned}
\end{equation}
This excludes locations within distance $d_{rival}$ of existing rival stations or within distance $d_{operator}$ of existing operator stations, addressing the spatial constraints observed in the current distribution of charging stations shown in Fig.\ref{fig:maps}.

The optimization is subject to the following constraints:
\begin{align}
    \sum_{v \in C} x_v &\leq M, \\
    (x_u + x_v) \cdot a_{uv} &\leq 1 \quad \forall\;(u,v) \in C, \\
    x_v & \in \{0,1\} \quad \forall\; v \in C,
\end{align}
where $M$ represents the maximum number of new stations to be installed, and $a_{uv}$ is an indicator equal to 1 if nodes $u$ and $v$ are within distance $d_{operator}$ of each other.

The second constraint ensures appropriate spatial distribution of stations, preventing clustering of new facilities and aligning with findings\cite{Wolbertus2021} suggesting that distributed charging networks offer superior service quality compared to concentrated hubs.

\subsection{Evaluation Metrics for Model Validity}

To ensure the robustness of the proposed methodology, both structural learning and optimization models undergo quantitative evaluation. The accuracy of the DAG-based demand estimation is assessed using the Structural Hamming Distance (SHD), which measures deviations between inferred and expected relationships in the data. Lower SHD values indicate higher structural accuracy. 
Fig.\ref{fig:SHD_vs_R} relates SHD to edge reversals, helping distinguish between structural errors and directional uncertainties in our causal models. 
We also use the F1-score as a standard information retrieval metric to evaluate edge prediction performance. These metrics are calculated for both raw and standardized data to assess the impact of feature scaling on structural learning performance. 

For the optimization model, two key performance indicators are used: demand satisfaction rate and coverage efficiency. The demand satisfaction rate quantifies the proportion of observed charging demand covered by the optimized station placement, ensuring that selected sites effectively serve user needs. Coverage efficiency assesses how much demand is captured per new station added, ensuring that infrastructure expansion follows an efficient, demand-driven approach.

These metrics provide a rigorous validation framework, ensuring that both causal inference and optimization contribute meaningfully to EV charging infrastructure planning.

\section{Results And Discussion}\label{sec:res}

This section presents our findings and their implications for EV charging infrastructure placement. We begin by examining empirical patterns in charging demand, followed by structural learning insights that reveal causal relationships. We then present optimization results for candidate charging station locations and discuss broader implications for infrastructure planning.

\subsection{Empirical Analysis of Charging Demand}

Analysis of charging data from Palo Alto and Boulder reveals distinct spatial patterns in station usage. As shown in Fig.\ref{fig:maps}, consumption levels vary significantly across locations, with higher demand (indicated in green) concentrated in specific areas. These patterns suggest that geographical features and surrounding amenities play crucial roles in station utilization.

\begin{figure}[htbp]
    \includegraphics[width=\linewidth]{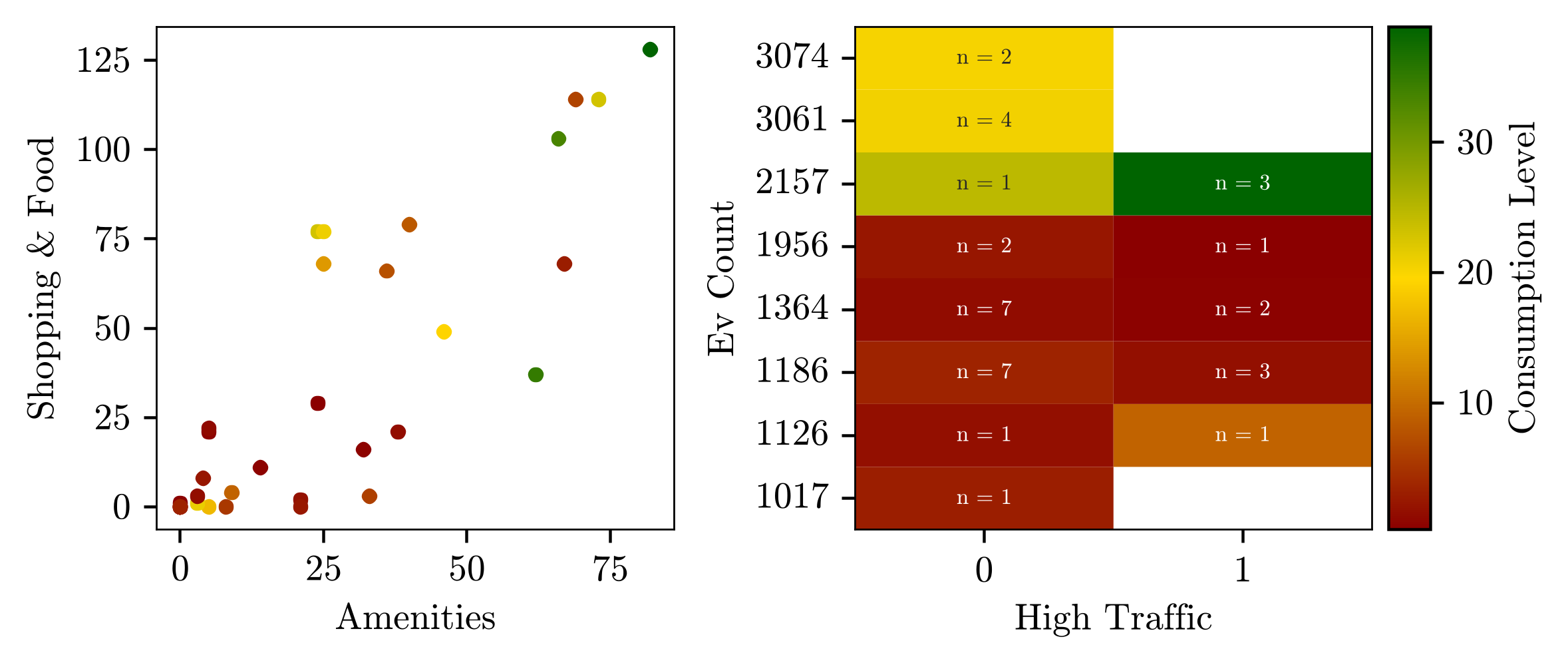}
    \caption{\textit{consumption level} by \textit{amenities} vs. \textit{shopping and food} (left) and \textit{consumption level} by \textit{high traffic} vs. \textit{ev count} (right)}
    \label{fig:link_plot_2}
\end{figure}

The data reveal several significant relationships between charging demand and contextual variables. Particularly noteworthy is the strong association between consumption levels and both amenity density and EV registration counts, as visualized in Fig.\ref{fig:link_plot_2}. Locations with higher densities of amenities, shopping, and food establishments consistently demonstrate higher charging consumption, suggesting that charging behavior is integrated with other activities rather than occurring in isolation. Similarly, proximity to high-traffic arterial roads correlates with increased utilization, especially in areas with higher EV registrations.

The comparative analysis between Palo Alto and Boulder data sets provides valuable insights into how charging infrastructure utilization evolves with different levels of EV adoption. With 48 charging stations across 10 locations in Palo Alto versus 52 stations across 25 locations in Boulder, the spatial distribution strategies differ notably between these municipalities despite similar total station counts.

\subsection{Structural Learning Insights}
\begin{figure}[htbp]
    \includegraphics[width=\linewidth]{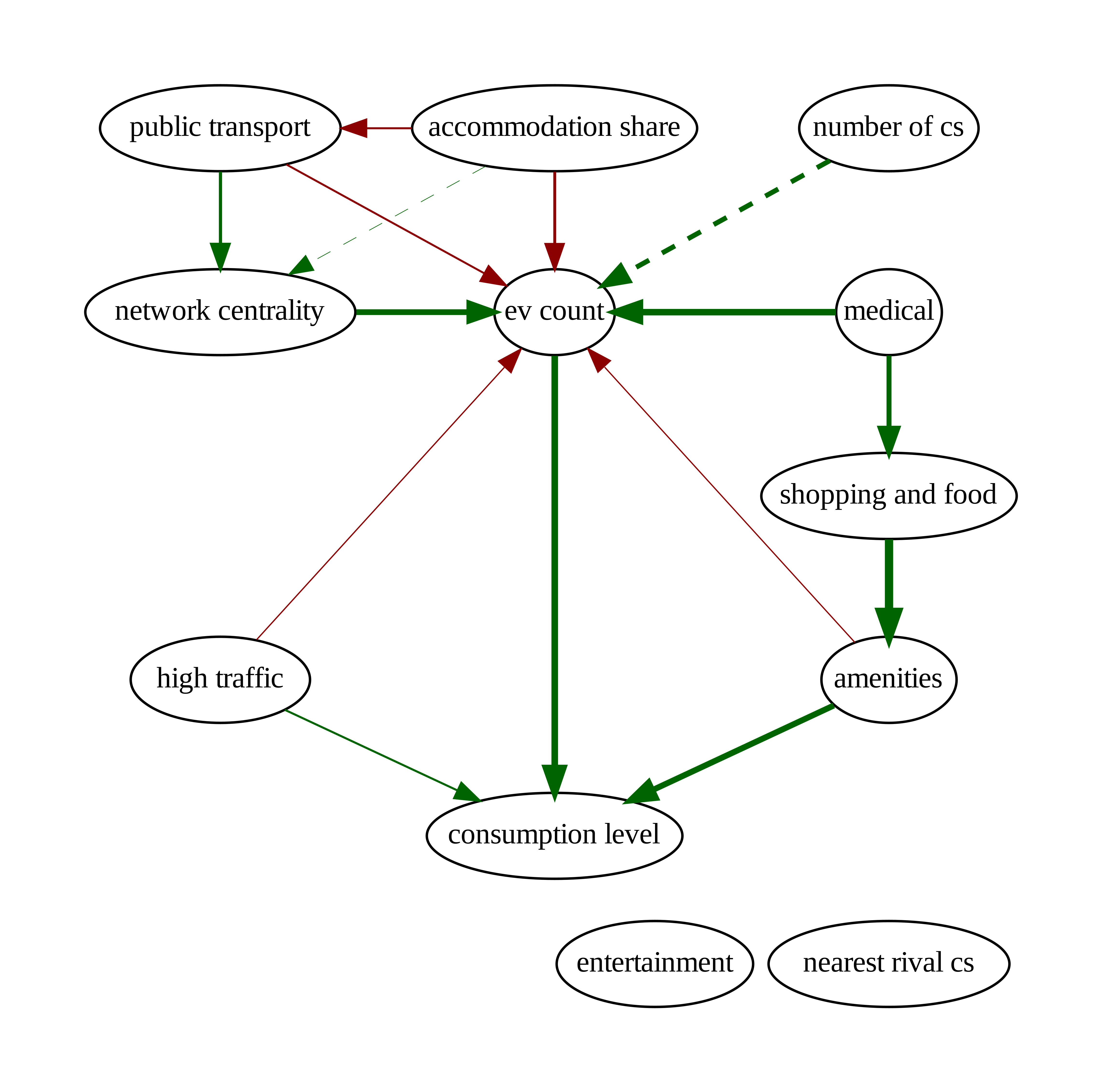}
    \caption{Reversed DAG as predicted by NOTEARS and DAGMA. Dashed edges were predicted by DAGMA, but not by NOTEARS. Green edges represent positive edge weights, and red edges represent negative edge weights. The edge thickness reflects the absolute value of the pairwise Pearson correlation coefficient between the nodes. For reference: weakest correlation: $\rho_{\textit{accommodation share, network centrality}} = -0.0623$; median correlation: $\rho_{\textit{public transport, network centrality}} = 0.3282$; strongest correlation: $\rho_{\textit{shopping and food, amenities}} = 0.8446$. Both models do not find any edges between \textit{entertainment} and \textit{nearest rival cs} and any of the other variables.}
    \label{fig:DAG}
\end{figure}

Our application of DAG-based structural learning yielded remarkably consistent results between the NOTEARS and DAGMA algorithms, suggesting robust statistical relationships in the underlying data. As visualized in Fig.\ref{fig:DAG}, both algorithms identified three primary determinants of charging consumption levels: proximity to amenities, EV registration density, and traffic flow. The consistency between these independently derived structures enhances confidence in the identified relationships.

The statistical validation of these structural insights through conditional independence tests confirms the significance of these relationships. The KCI test results presented in Table \ref{tab:KCIresults} show that variables such as public transport accessibility and network centrality may have marginal conditional relationships with consumption levels (p-values of 0.0368 and 0.0600, respectively), while accommodation share, number of charging stations at a location, medical facilities, and shopping and food establishments demonstrate clear conditional independence (p-values ranging from 0.4123 to 0.9742).

\begin{table}[htbp]
    \caption{Mean and standard deviation of p-values from KCI test. Each row tests whether consumption level is conditionally independent of the listed variable, given amenities, EV count, and high traffic proximity.}
    \label{tab:KCIresults}
    \centering
    \begin{tabularx}{\linewidth}{Xrr}
    \toprule
    \textbf{Conditional Independence Hypothesis} & \textbf{Mean} & \textbf{SD} \\
    \midrule
    Public transport & 0.0368 & 0.0030 \\
    Accommodation share & 0.4123 & 0.0075 \\
    Number of CS & 0.5669 & 0.0059 \\
    Network centrality & 0.0600 & 0.0039 \\
    Medical facilities & 0.9742 & 0.0024 \\
    Shopping and food & 0.6916 & 0.0056 \\
    \bottomrule
    \end{tabularx}
\end{table}

\begin{figure}[htbp]
    \includegraphics[width=\linewidth]{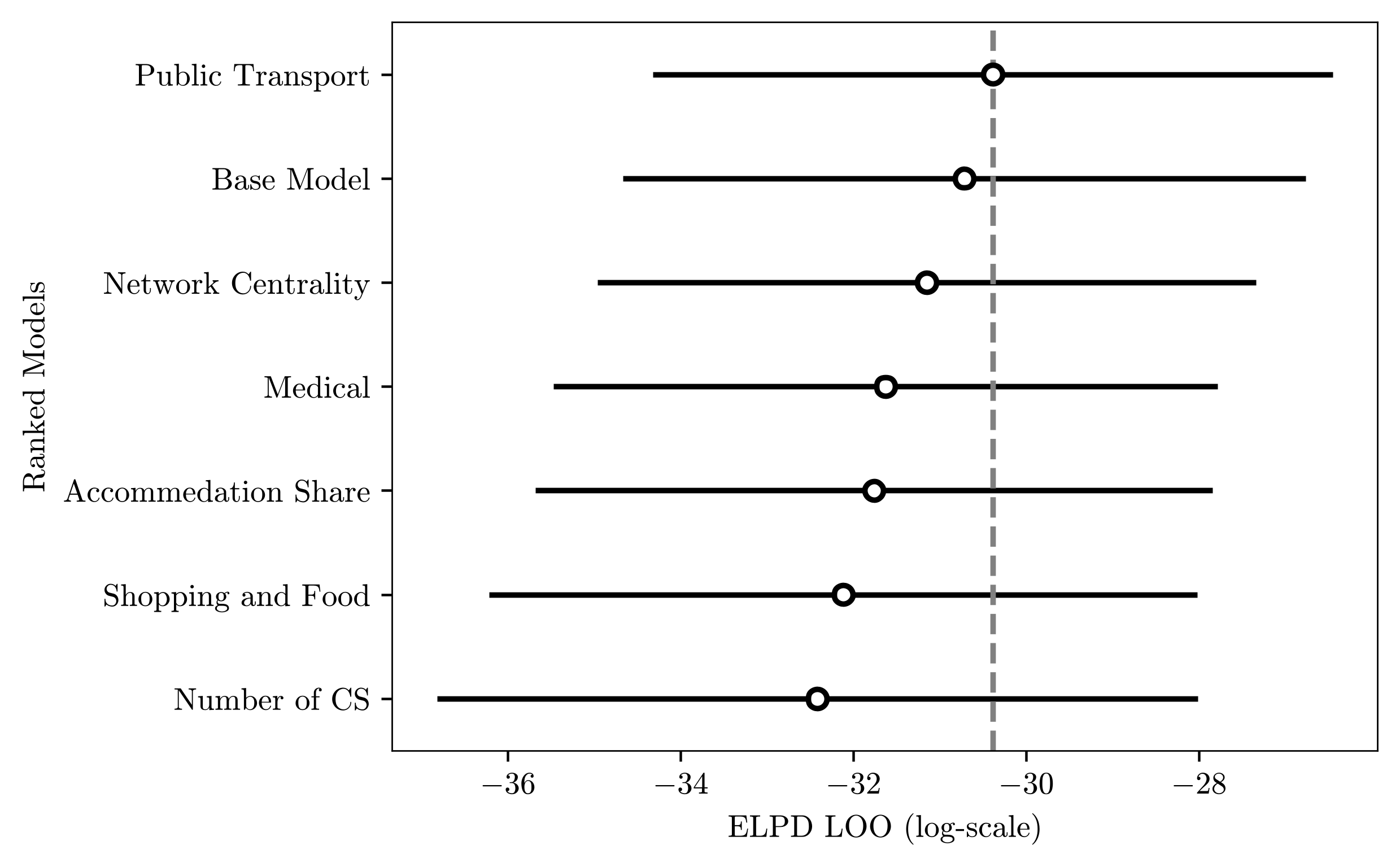}
    \caption{ELPD of different regression models}
    \label{fig:model_comparison_loo1}
\end{figure}

Bayesian regression models further validate these findings, with the model comparison illustrated in Fig.\ref{fig:model_comparison_loo1} indicating that the baseline model incorporating only amenities, EV counts, and high-traffic proximity provides the most parsimonious explanation of charging demand. The expected log predictive density (ELPD) values demonstrate that adding additional variables does not significantly improve predictive performance.

These results challenge certain assumptions in the existing literature. Notably, our findings suggest that the number of charging stations at a location and the distance to rival charging stations do not directly influence consumption levels when controlling for amenity density, EV registrations, and traffic proximity—a finding that contradicts previous research positing direct competitive effects among charging stations.

\subsection{Optimization Outcomes And Recommended Sites}

\begin{figure}[htbp]
    \includegraphics[width=\linewidth]{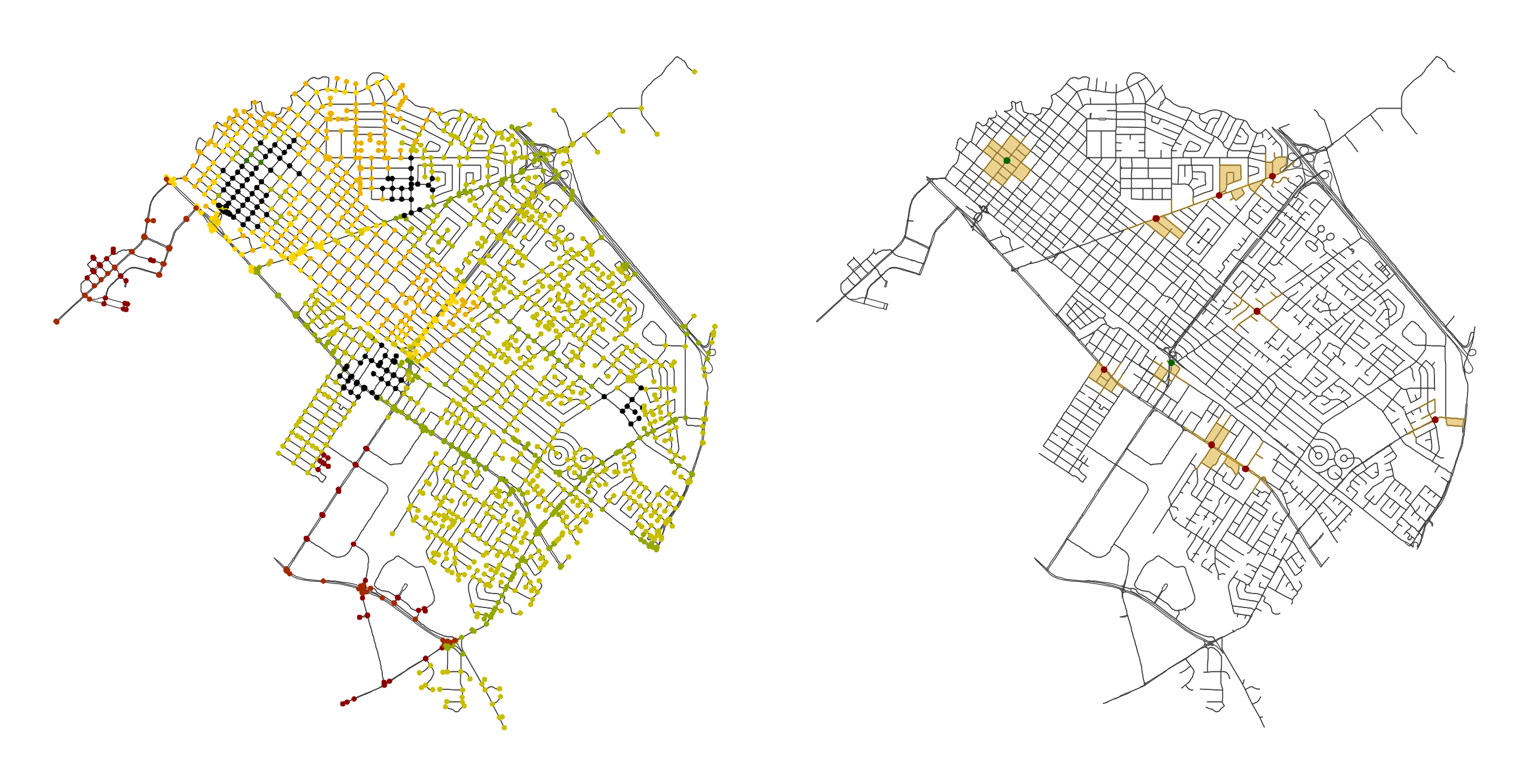}
    \caption{Scores $s_v \; \forall \; v \in C$ (left) and selected candidates with respective \textit{walkable areas} (right).}
    \label{fig:optimization_results}
\end{figure}

Based on our structural learning insights, we formulated an optimization model that prioritizes locations with high amenity density, substantial EV registrations, and proximity to high-traffic roads. The optimization results, displayed in Fig.\ref{fig:optimization_results}, identify ten candidate locations for new charging stations in Palo Alto, with two high-priority locations highlighted in green.

\begin{figure}[htbp]
    \includegraphics[width=\linewidth]{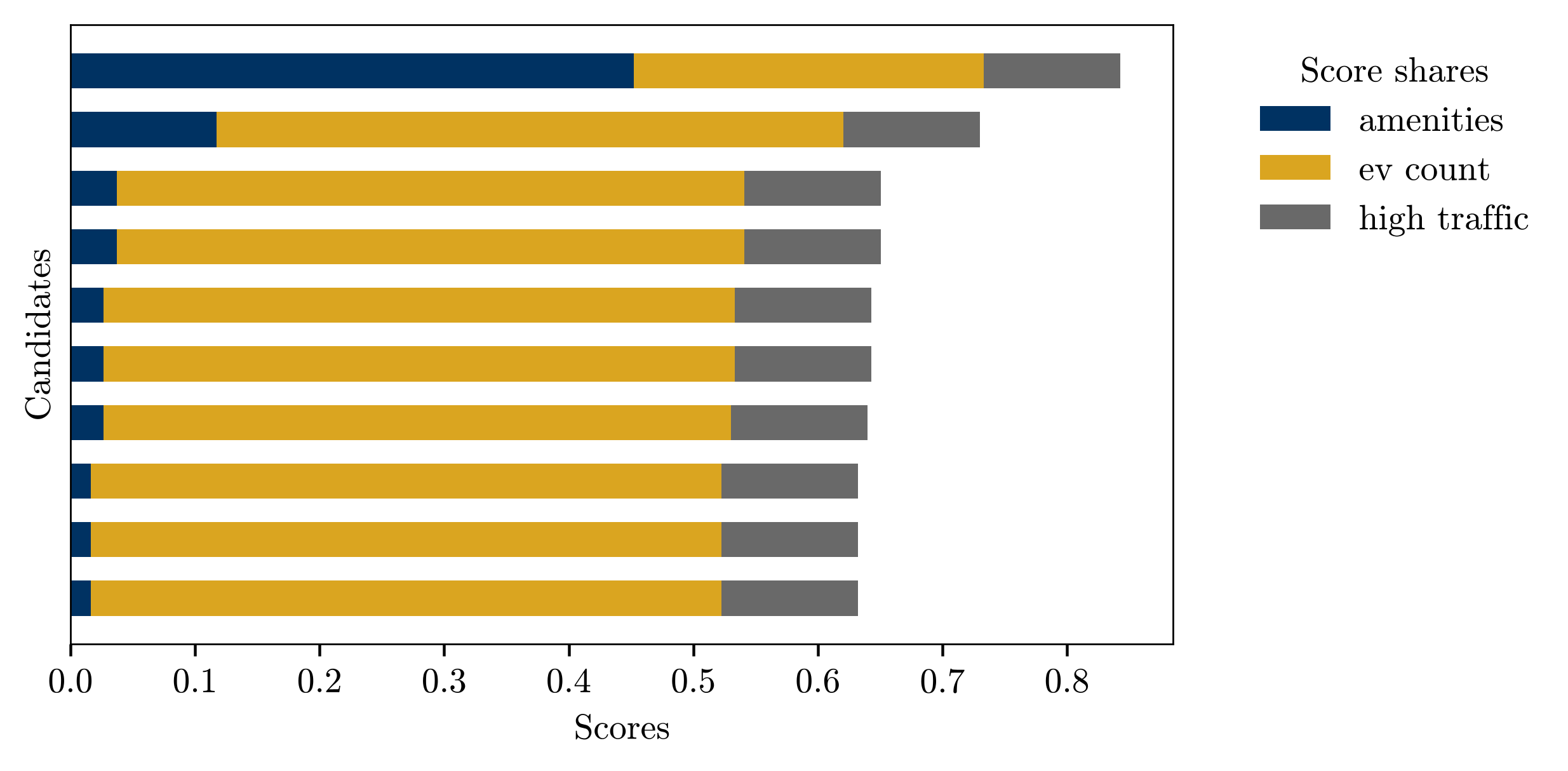}
    \caption{Scores of the ten candidate locations broken down into the contributions of each feature.}
    \label{fig:disaggregated_scores}
\end{figure}

A disaggregated analysis of these candidate locations, presented in Fig.\ref{fig:disaggregated_scores}, reveals that the three key features differentially influence their scores. While the highest-scoring candidate locations benefit from balanced contributions across all features, locations 3-10 show a heavier reliance on EV registration counts and high-traffic proximity, with relatively lower amenity contributions.

The spatial distribution of these candidate locations shows an interesting pattern: they tend to cluster in areas with existing charging infrastructure but with sufficient distance to avoid cannibalization. This suggests that the optimal strategy for Palo Alto may involve densifying charging infrastructure in already-established EV-friendly zones rather than expanding into entirely new areas—a finding consistent with the city's advanced stage of EV adoption.

\subsection{Implications for Infrastructure Planning}

Our findings have several important implications for infrastructure planners and policymakers. First, they confirm that the relationship between charging infrastructure and EV adoption may be bidirectional: areas with higher EV registrations benefit from additional charging stations, which may further accelerate adoption. This creates a positive feedback loop that infrastructure planners should consider when allocating resources.

Second, the importance of amenity density in determining charging station usage suggests that integration with existing commercial and service infrastructure should be a priority. Charging stations placed near diverse amenities enable productive use of charging time and increase overall station utilization—a win-win scenario for both users and operators.

\begin{figure}[htbp]
    \centering
    \includegraphics[scale=0.8]{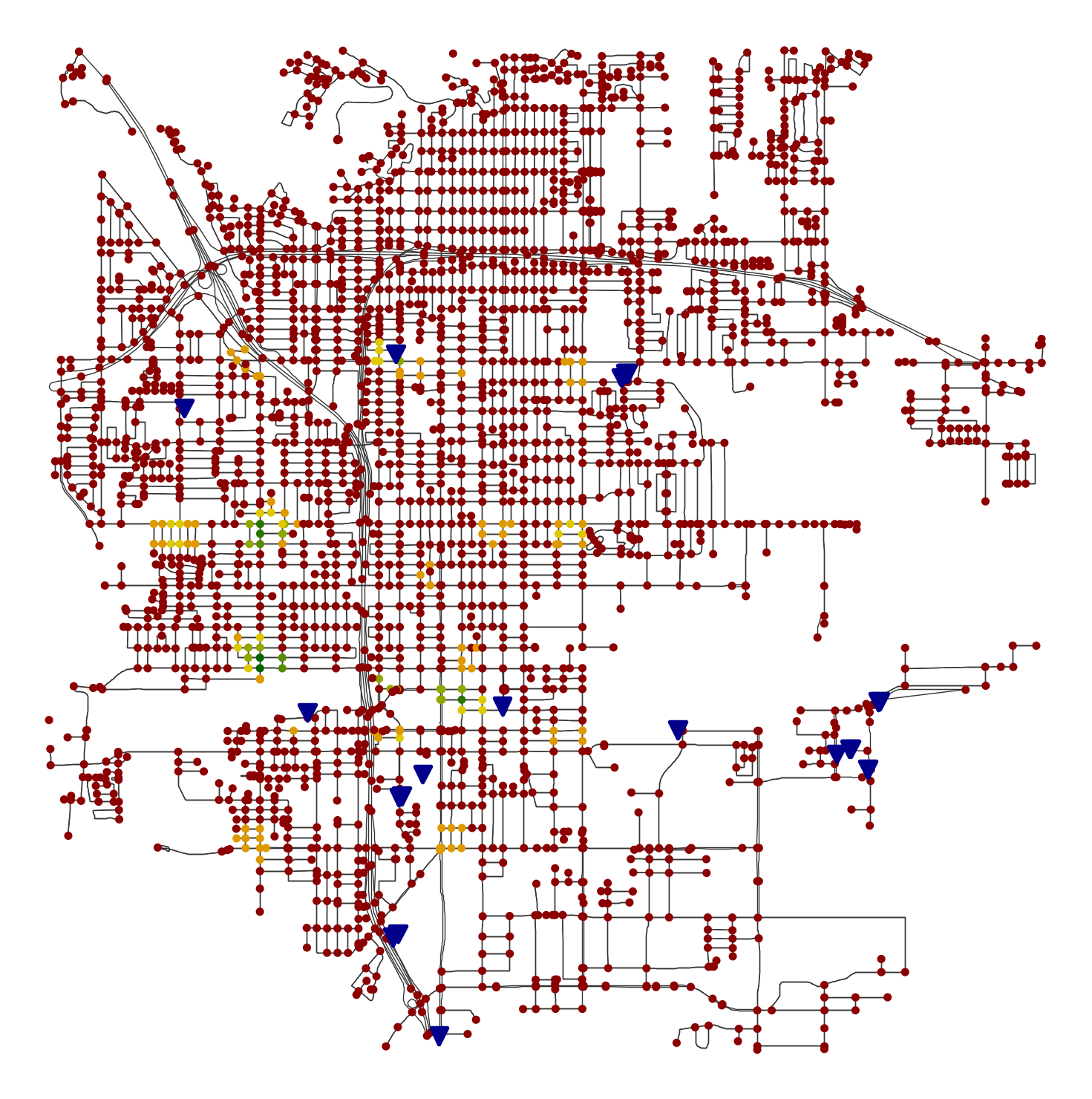}
    \caption{San Bernardino nodes colored according to the logarithm of \textit{amenities} plus \textit{shopping and food} in their \textit{walkable area} with CSs (blue triangles).}
    \label{fig:san_bernardino_scores}
\end{figure}

Third, our comparison with San Bernardino (Fig.\ref{fig:san_bernardino_scores}) highlights how optimization priorities may differ based on a city's stage of EV adoption. While Palo Alto's advanced adoption stage suggests a strategy of targeted densification in high-demand areas, cities with earlier adoption patterns, like San Bernardino, may benefit more from broader geographical coverage to stimulate adoption.

Finally, these findings challenge the common assumption that charging stations should be distributed evenly across a city. Our analysis suggests that strategic clustering in high-amenity, high-traffic areas with substantial EV populations may be more effective for optimizing infrastructure utilization, particularly in cities with advanced EV adoption. However, this approach must be balanced against equity considerations to ensure adequate access for all residents.

The methodology presented here offers a transferable framework for cities at various stages of EV adoption. By combining empirical data analysis with structural learning and optimization techniques, planners can develop evidence-based strategies for charging infrastructure deployment that respond to their communities' specific characteristics and needs.

\section{CONCLUSION}\label{sec:con}

This paper presented a data-driven approach for optimizing electric vehicle charging station placement by applying causal discovery techniques to empirical charging data. Analysis of charging patterns in Palo Alto and Boulder revealed that station utilization is primarily determined by three factors: proximity to amenities, EV registration density, and adjacency to high-traffic routes. These findings, consistent across both urban contexts and multiple algorithms, challenge conventional infrastructure distribution strategies.

The optimization framework we developed translates these insights into actionable placement recommendations, identifying locations likely to experience high utilization based on the discovered dependency structures. This approach aligns charging infrastructure with actual usage patterns rather than theoretical assumptions, potentially enhancing both station utilization and user convenience.

Despite limitations in generalizability to emerging markets and challenges in definitive causal inference, this research contributes a methodologically rigorous framework that integrates empirical charging behavior into infrastructure planning decisions. By prioritizing data-driven insights over theoretical distribution models, this approach offers a more efficient strategy for expanding charging networks that can adapt to different stages of EV market development.
The integration of structural learning with optimization techniques establishes a novel approach for infrastructure planning that can accelerate the transition to electrified transportation through more effective charging network design.
\bibliographystyle{ieeetran}
\bibliography{Bibliography.bib}

\end{document}